\documentclass[final]{cvpr}

\usepackage{times}
\usepackage{epsfig}
\usepackage{graphicx}
\usepackage{amsmath}
\usepackage{amssymb}

\usepackage[pagebackref=true,breaklinks=true,colorlinks,bookmarks=false]{hyperref}

\def\cvprPaperID{****} 
\def\confYear{CVPR 2021}

\begin{document}

\title{TANet++: Triple Attention Network with Filtered Pointcloud on 3D Detection}

\author{Cong Ma\\
Peking University\\
Beijing, China\\
{\tt\small Cong-Reeshard.Ma@pku.edu.cn}
}

\maketitle

\begin{abstract}
TANet~\cite{liu2020tanet} is one of state-of-the-art 3D object detection method on KITTI~\cite{Geiger2013IJRR} and JRDB~\cite{martin2021jrdb}  benchmark, the network contains a Triple Attention module and Coarse-to-Fine Regression module to improve the robustness and accuracy of 3D Detection. However, since the original input data (point clouds) contains a lot of noise during collecting the data, which will further affect the training of the model. For example, the object is far from the robot, the sensor is difficult to obtain enough pointcloud. If the objects only contains few point clouds, and the samples are fed into model with the normal samples together during training, the detector will be difficult to distinguish the individual with few pointcloud belong to object or background. In this paper, we propose TANet++ to improve the performance on 3D Detection, which adopt a novel training strategy on training the TANet. In order to reduce the negative impact by the weak samples, the training strategy previously filtered the training data, and then the TANet++ is trained by the rest of data. The experimental results shows that AP score of TANet++ is 8.98\% higher than TANet on JRDB benchmark.
\end{abstract}

\section{Data Filter Strategy}

TANet++ refers to TANet structure (one of the state-of-the-art 3D detection method) as 3D detection model with same hyper-parameters and training strategy. To reduce the impact of the bad samples for training the TANet, we propose three constraints to filter the bad samples and ensure the vaildity of the training data. 

The constraints includes: 1.Distance Constraint; 2.Number of Pointcloud Constraint; 3.Occlusion Constraint.

\subsection{Distance Constraint}

Since the objects are too far away from the sensor, the captured data does not contain enough point cloud information to support the representation of the target features. Furthermore, the model cannot distinguish these objects and background well, which cause the output of model might contain a large number of false positives (background are mis-detected as specific object). Therefore, Distance Constraint aims to filter the objects which is far from the sensor, the constraint is shown as :

\begin{eqnarray}
	\left\{\begin{matrix}
		discard,\  {||p_s-p_i ||}_2^2> \eta\\
		keep,\  {||p_s-p_j ||}_2^2\leq \eta\\
	\end{matrix}\right.
\end{eqnarray} 

where $p_s$ and $p_i$ indicate the center of 3D position of the sensor and object respectively, $\eta$ is distance threshold, which is set as 15. If the distance between sendor and object $i$ is less than $\eta$, then the object $j$ will be kept in the training set, other wise the object will be remove from the training set.

\subsection{Number of Pointcloud Constraint}

Since occlusion and the ladar interference, some targets only contain few pointcloud in their bounding boxes. Even though the object is close to the senor, the number of pointcould in the target is still limited if the target is seriously occluded. Therefore, we propose Number of Pointcloud Constraint to remove the targets which do not contain enough pointcloud, the constraint is shown as:

\begin{eqnarray}
	\left\{\begin{matrix}
		discard,\  n_i < \delta\\
		keep,\  n_j \geq \delta\\
	\end{matrix}\right.
\end{eqnarray} 

where $n_i$ and $n_j$ indicates the number of pointcloud of object $i$ and $j$ respectively. $\delta$ is the number threshold, which is set as 10. If the number of pointcloud of target is more than $delta$, then the object $j$ will be kept in the training set, other wise the object will be remove from the training set.

\begin{figure*}
	\centering
	\includegraphics[width=16cm]{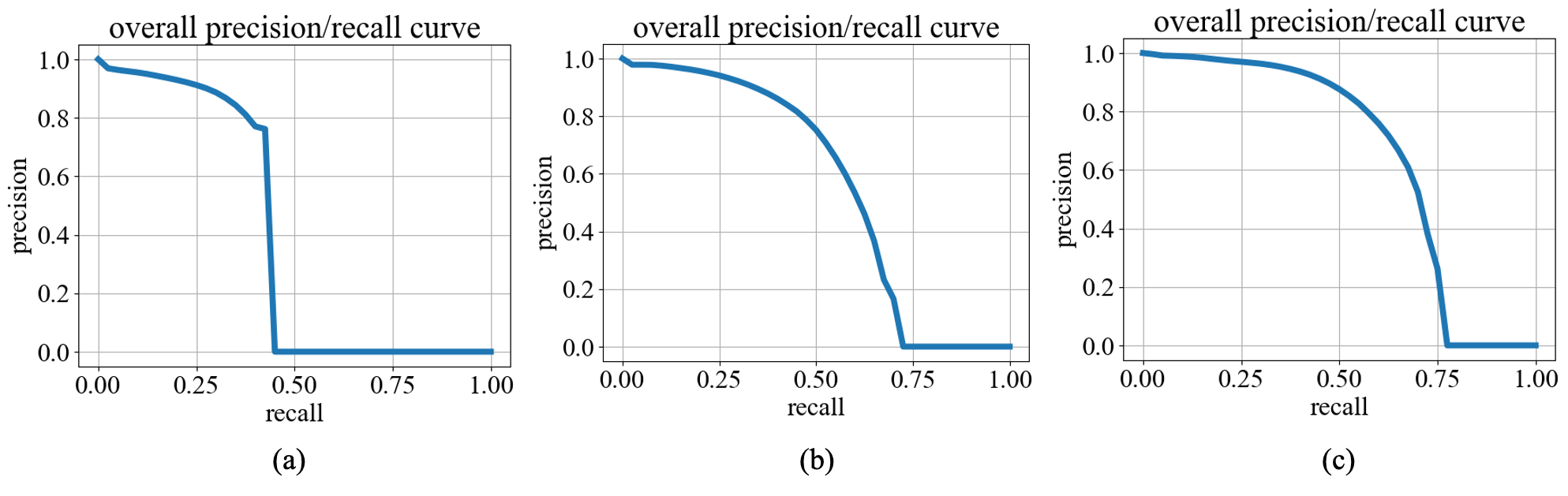}
	\caption{Overall Precision-Recall curves: (a): F-PointNet (b):TANet (c):TANet++}
	\label{fig3}
\end{figure*}

\begin{figure*}
	\centering
	\includegraphics[width=16cm]{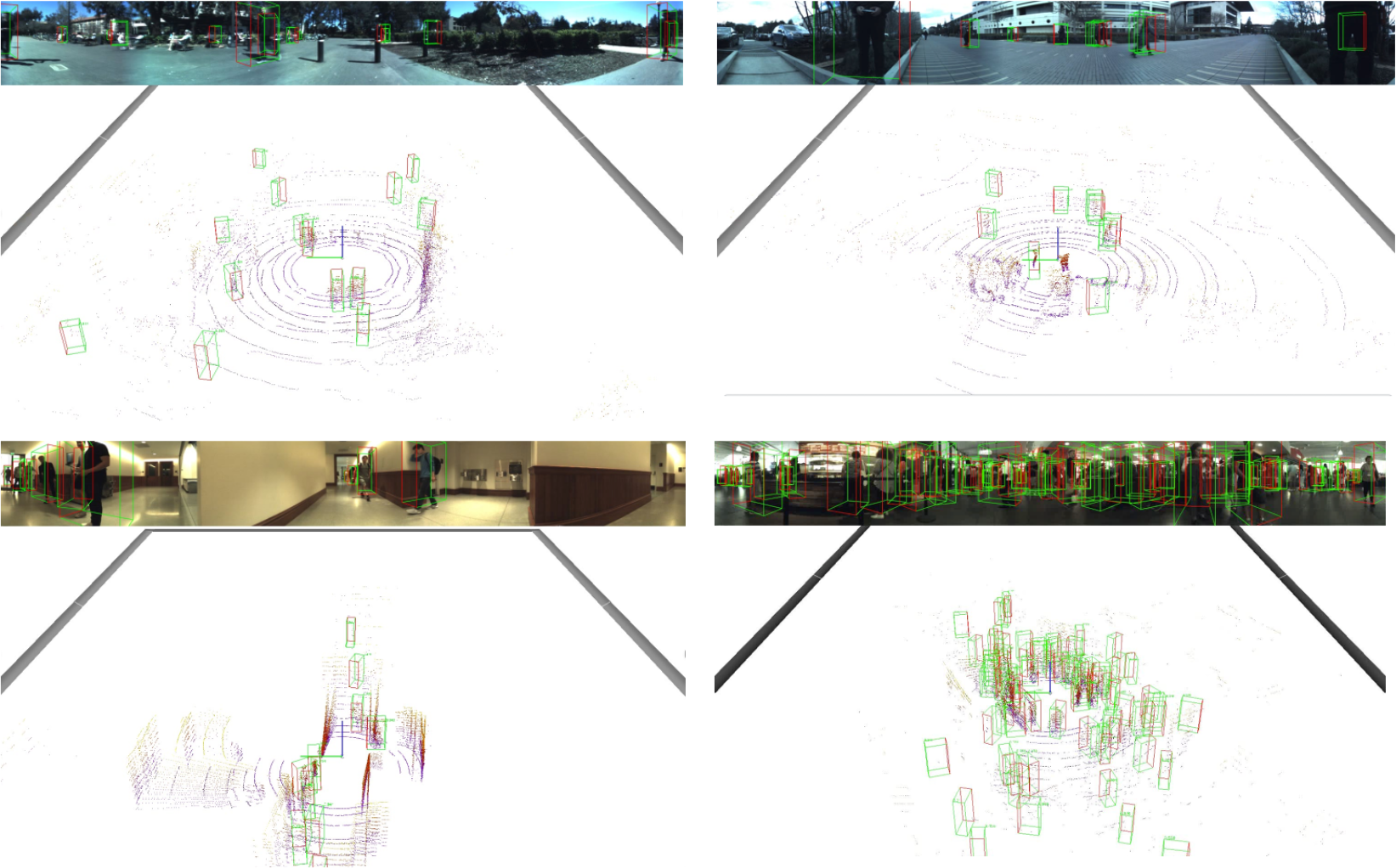}
	\caption{Overall Precision-Recall curves: (a): F-PointNet (b):TANet (c):TANet++}
	\label{fig4}
\end{figure*}

\subsection{Occlusion Constraint}
JRDB~\cite{martin2021jrdb} additionally provides the information that the annotated target is occluded or not, which includes "Fully Visible", "Mostly Visible", "Severly Occluded" and "Fully Occluded". These information can help us further filter the bad samples. The constraint is shown as:

\begin{eqnarray}
	\left\{\begin{matrix}
		discard,\  i \in "Fully Occluded"\\
		keep,\  j \in others\\
	\end{matrix}\right.
\end{eqnarray} 

where if the object $i$ is labelled as "Fully Occluded", then the target will be remove from the training set, otherwise the target will be kept in the training set.

\section{Experimental Results} 

\begin{table}[h]
	\begin{center}
		\begin{tabular}{l|c}
			\hline
			Method & AP \\
			\hline
			F-PointNet~\cite{qi2018frustum} & 38.21\\
			TANet~\cite{liu2020tanet} & 54.94 \\
			TANet++ & 63.92\\
			\hline
		\end{tabular}
	\end{center}
	\caption{Results on JRDB testing set}
\end{table}

Table 1 shows the detection results on JRDB testing set, where our method TANet++ is 8.98\% and 25.71\% higher than TANet and F-PointNet respectively on AP.

Figure 1 illstrates the precision-recall curves for F-PointNet, TANet and TANet++.

Figure2 shows the TANet++ visulization results in JRDB testing set.

{\small
\bibliographystyle{ieee_fullname}
\bibliography{egbib}
}

\end{document}